\documentclass{article} 
\usepackage{colm2024_conference}

\usepackage{microtype}
\usepackage{hyperref}
\usepackage{url}
\usepackage{booktabs}
\usepackage{diagbox}
\usepackage{makecell}
\usepackage{colortbl}
\usepackage{graphicx}
\usepackage{longtable}
\usepackage{subcaption}

\definecolor{darkblue}{rgb}{0, 0, 0.5}
\hypersetup{colorlinks=true, citecolor=darkblue, linkcolor=darkblue, urlcolor=darkblue}

\colmfinalcopy 

\title{SambaLingo: Teaching Large Language Models New Languages}

\author{Zoltan Csaki, Bo Li, Jonathan Li, Qiantong Xu, Pian Pawakapan, Leon Zhang, Yun Du,\\ \textbf{Hengyu Zhao,  Changran Hu \& Urmish Thakker}\\
SambaNova Systems\\
\texttt{zoltan.csaki@sambanovasystems.com}
}

\begin{document}

\maketitle

\begin{abstract}
Despite the widespread availability of LLMs, there remains a substantial gap in their capabilities and availability across diverse languages. One approach to address these issues has been to take an existing pre-trained LLM and continue to train it on new languages. While prior works have experimented with language adaptation, many questions around best practices and methodology have not been covered. In this paper, we present a comprehensive investigation into the best practices for adapting LLMs to new languages. Our study explores the key components in this process, including vocabulary extension and initialization of new tokens, direct preference optimization and the data scarcity problem for human alignment in low-resource languages. We scale these experiments across 9
languages and 2 parameter scales (7B and 70B). We compare our models
against Llama 2, Aya-101, XGLM, BLOOM and existing language experts,
outperforming all prior published baselines. Additionally, all evaluation code\footnote{\label{foot1}Fork of lm-evaluation-harness \citep{eval-harness} with new multilingual benchmarks: \url{https://github.com/sambanova/lm-evaluation-harness}} and checkpoints\footnote{\label{foot2}All SambaLingo Checkpoints: \url{https://huggingface.co/collections/sambanovasystems/sambalingo-65e25770f2037c85ad35ca77}} are made public to facilitate future research.
\end{abstract}
\section{Introduction}
\label{Introduction}
New state of the art large language models are being released at a breakneck speed, yet their training data, tokenizer, and evaluations remain primarily centered around a few popular languages such as English, Chinese, French and Arabic. In principle, the way to create large language models for specific languages is to pre-train models from scratch \citep{sengupta2023jais, zhang2020cpm}. However, it is difficult to obtain a large amount of compute resources and a vast quantity of data in diverse languages. Researchers have tackled this problem by training monolithic multi-lingual models that cover a wide range of languages \citep{workshop2023bloom, lin2022fewshot, shliazhko2023mgpt, xue2021mt5}. These models can still struggle to achieve uniformly good results across all languages due to various factors such as the curse of multilinguality \citep{chang2023multilinguality, conneau2020unsupervised} and the scarcity of pre-training data in many languages \citep{chung2023unimax}.  

\begin{figure}
\centering
\includegraphics[width=14cm]{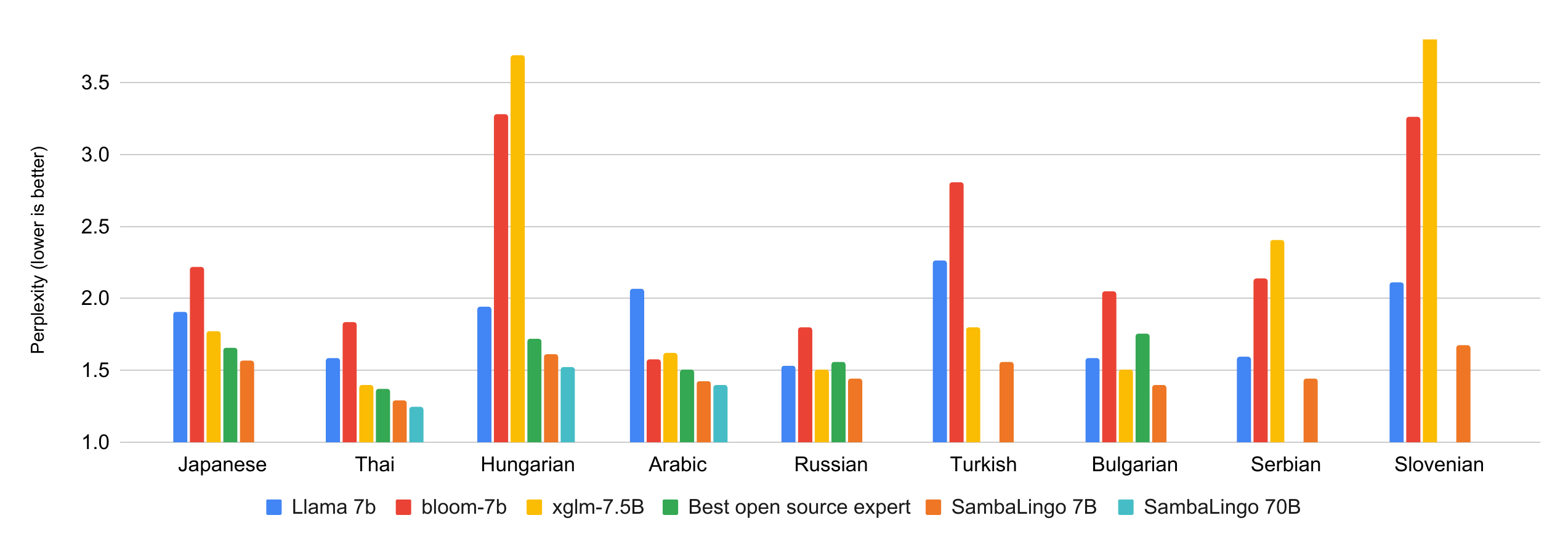}
    \caption{Evaluation perplexity on hold out dataset, we also evaluate perplexity over wikipedia and Mc4 in appendix \ref{MRT}. Open source expert baselines: Japanese - Swallow-7b-hf \citep{swallow}, Thai: typhoon-7b \citep{pipatanakul2023typhoon},  Arabic: jais-13b \citep{sengupta2023jais}, Hungarian: PULI-GPTrio \citep{yang-puli-gptrio}, Russian: saiga-7b \citep{IlyaGusevsaiga}, Bulgarian: mGPT-bulgarian\citep{shliazhko2023mgpt}. We could not find Serbian, Slovenian and Turkish languages models with low enough perplexity that would fit the graph so we chose to omit them here to ensure readability. 
    }
    \label{fig:perplexity_results}
\end{figure}

Recently, adapting English centric models to new languages has gained prominence \citep{blevins2024breaking, yong2023bloom1, ebrahimi2021adapt, Pires_2023, pipatanakul2023typhoon, lin2024mala500}. The resulting models can outperform large multilingual models and even language specific models pre-trained from scratch. Adaptation requires various design choices around the tokenizer, data, alignment and evaluation strategies. This paper aims to provide a comprehensive study to help inform these decisions, outlining a protocol to adapt a pre-trained model to a new language. We show that our methodology works by training models across 9 languages and 2 parameter scales (7B and 70B). Figure \ref{fig:perplexity_results} and \ref{fig:Main_Results} show that our methodology can lead to better models than existing state of the art models.
\\
\\
The key studies and contributions include:
\begin{itemize}
  \item Best practices for adapting existing LLMs to new languages scaled across 9 typologically and linguistically diverse languages including Arabic, Bulgarian, Hungarian, Japanese, Russian, Serbian, Slovenian, Thai, and Turkish
  \begin{itemize}
            \item{Expanding the vocabulary for the target language improves the tokenizer fertility (\ref{fig:token_fertility}), but does not have a siginficant impact on downstream accuracy (\ref{Vocabulary Expansion vs Original Tokenizer})} 
            \item{Various embedding initialization methods have minimal impact on accuracy, but sub word averaging accelerates training loss convergence (\ref{New Token Embedding Initialization})}
            \item{The quality of the base checkpoint on English benchmarks can improve downstream language adaptation results (\ref{Importance Of Base Model Quality})}
  \end{itemize}   
  \item A recipe for human preference alignment in any language using open source data
  \begin{itemize}
            \item{Aligning the adapted model requires minimal data from the target language, reducing the need of gathering expensive alignment data (\ref{DPO Data Mixture})}
            \item{The choice of translated versus human written alignment data does not have a large impact on win rates (\ref{abl:dpo})}
  \end{itemize} 
  \item Open sourcing code and checkpoints to promote future research
  \begin{itemize}
            \item{State of the art models adapted from Llama 2 in 9 languages and 2 parameter scales (7B, 70B)\footref{foot2}}
            \item{Integration of FLORES-200, SIB-200, EXAMS and multilingual perplexity benchmarks with lm-eval-harness\footref{foot1} \citep{eval-harness}}
  \end{itemize} 
\end{itemize}

\section{Related Work}
\label{relWork}
While prior work has explored adapting pre-trained LLMs to new languages, they do not extensively study the methodology to do so. None of these works explore the design choices around aligning models in new languages, for example the mixture of data in the base models language and the new language or the impact of translated data on qualitative evaluation. \citet{Pires_2023} and \citet{cui2023efficient} adapt Llama models to Portuguese and Chinese respectively, but they do not explore the impact of vocabulary extension and/or initialization. \citet{blevins2024breaking} explores training language experts to break the curse of multilinguality starting from a pre-trained model, but they do not explore the impact of vocabulary extension, initialization and quality of the base model. Extension of vocabulary was discussed in \citet{zhao2024llama,tikhomirov2023impact}, however they do not explore token embedding initialization strategies or impact of quality of base model. \citet{lin2024mala500} studies simultaneous language adaptation to 500 languages. Nevertheless, they also do not answer questions around alignment or token initialization strategy. \citet{ye2023language} studies language adaptation of a wide variety of English-centric and multilingual models, however they only focus on fine-tuning XNLI tasks.

There has been a significant body of work around open-source multi-lingual models \citep{workshop2023bloom, lin2022fewshot, shliazhko2023mgpt}. Our work differs from the aforementioned studies as we solely focus on adapting pre-trained LLMs to new languages and not on pre-training from scratch. Notably, these multilingual open-source models tend to be pretrained on significantly fewer tokens than the base models we adapt from. As the models in this work tend to outperform these multilingual models, this presents a promising path forward for obtaining the state of the art in new languages.

\section{Adaptation Methodology}
\label{Adaptation Methodology}
We present our methodology to adapt large languages models to a new language, with state of the art results in 9 target languages: Arabic, Thai, Turkish, Japanese, Hungarian, Russian, Bulgarian, Serbian and Slovenian. We select these languages because they provide a mix of high resource and lower resources languages with diverse character sets and linguistic patterns. We additionally limit the scope of the languages studied in this paper to languages with easily available text datasets from CulturaX \citep{nguyen2023culturax}. See Section \ref{Evaluation} for evaluation results on the final checkpoints produced by this methodology, and Section \ref{Ablations} for ablations justifying our methods.

We use the term \textit{initial language} to describe the original language that the base model was trained on (in this case, English) and the term \textit{target language} as the new language this model is being adapted to.

\subsection{Selecting a Base Model}\label{BaseCheckpoint}

Our methodology starts with an existing base checkpoint instead of pre-training from scratch. Previous work has shown that starting from an existing checkpoint leads to faster training convergence, better downstream evaluation accuracy and lower compute/data requirements \citep{Pires_2023, lin2024mala500, csaki2023efficiently}. Section \ref{Importance Of Base Model Quality} demonstrates that it is important to select a starting checkpoint with the best results for the initial language, as that will improve the downstream results for the target language. Based on these observations, we chose Llama2 7B as our base model to adapt to target languages, the best open source model available at the time of the experiments.

We additionally scale this methodology to Llama 2 70B. Given compute restrictions, we only do this for 3 languages - Arabic, Thai and Hungarian. See Section \ref{Scaling From 7B to 70B} for in-depth comparisons of our 7B and 70B models.

\subsection{Extending Model Vocabulary}

Llama 2 \citep{touvron2023llama} was trained predominantly on English text, and has poor tokenizer efficiency for other languages (see Section \ref{VocabularyExpansion}). To address this inefficiency, we chose to extend the vocabulary of the Llama 2 tokenizer by adding non overlapping tokens from the target language and initializing them using sub-word embeddings from the original tokenizer. See Section \ref{VocabularyExpansion} for experiments that justify our approach.

\subsection{Continual Pre-training}\label{Continuous Pretraining}
We train each language independently on data that consists of a 1:3 mixture of English and target language web data biased towards the target language. Pretraining data for all languages, including English, is sourced from CulturaX \citep{nguyen2023culturax}.
These decisions are grounded in results from previous works: \citet{zhao2024llama, csaki2023efficiently} show that mixing in data from the base model domain helps downstream accuracy and training stability, \citet{gupta2023continual} find that including a higher proportion of data from the target distribution helps improve the convergence in the target distribution, \citet{almazrouei2023falcon} showed the importance of cleaned web data. Additionally, hyperparameters used for training can be found in Appendix \ref{app:hyper}.

\subsection{Aligning To Human Preferences In Other Languages}\label{Aligning To Human Preferences In Other Languages}
To train a chat-aligned version of the model, we follow the two-stage approach from \citet{tunstall2023zephyr} - supervised finetuning (SFT) followed by direct preference optimization (DPO). 
More details about hyperparameters for each of these phases used can be found in Appendix \ref{app:hyper}.

\begin{itemize}
    \item For SFT, we use \texttt{ultrachat-200k} \citep{tunstall2023zephyr}, in a 1:1 ratio with a Google translated version of \texttt{ultrachat-200k}. 
    \item For human preference alignment, we use the \texttt{ultrafeedback} \citep{cui2023ultrafeedback} and \texttt{cai-conversation-harmless} dataset \citep{Huang2024cai}. We mix these datasets with a 10:1 ratio of English to machine translated data. Section \ref{DPO Data Mixture} shows that this ratio of data performs almost as well as other ratios and section \ref{abl:dpo} shows that machine-translated data can perform as well as human written data. 
\end{itemize}

\section{Evaluation}
\label{Evaluation}
\subsection{Quantitative Evaluation}\label{Quantitative Evaluation}
\begin{table}[h]
\centering
\begin{tabular}{|l|l|l|l|l|}
\hline
\rowcolor[HTML]{C0C0C0} 
Datasets &
  \begin{tabular}[c]{@{}l@{}}Task \\ Category\end{tabular} &
  \begin{tabular}[c]{@{}l@{}}Num \\ Few-Shot\end{tabular} &
  \begin{tabular}[c]{@{}l@{}}Number Of \\ Languages\end{tabular} &
  Metric \\ \hline
mc4 , Wikipedia & Perplexity  & \textbf{-} & 323 & Perplexity \\ \hline
FLORES-200      & Translation & 8           & 200           & CHRF          \\ \hline
SIB-200         & Text Classification & 3 & 200 & Accuracy \\ \hline
\cellcolor[HTML]{FFFFFF}BELEBELE & \cellcolor[HTML]{FFFFFF}Question Answering & 3 & 122 & Accuracy \\ \hline
Exams          & Knowledge   & 3           & 11            & Accuracy            \\ \hline
{\begin{tabular}[c]{@{}l@{}}XNLI\\ XStoryCloze\\  XCOPA \\ XWinograd \\ PAWS-X \end{tabular}} &
  {\begin{tabular}[c]{@{}l@{}}Natural Language \\ Understanding\end{tabular}} &
  {0} &
  {25+} &
  {Accuracy} \\ \hline
\end{tabular}
\caption{Multi-lingual evaluation suite}
\label{tab:eval_bench}
\end{table}
We use a wide variety of benchmarks to quantitatively evaluate the performance of our models and compare them to prior work. See Table \ref{tab:eval_bench} for the full list of quantitative benchmarks. In summary, we evaluate language modeling with perplexity on a holdout set of CulturaX \citep{nguyen2023culturax}, translation with CHRF \citep{popovic-2015-chrf} on FLORES-200 \citep{flores, zhu2023multilingual}, text classification accuracy on SIB-200 \citep{sib, lin2024mala500}, open-book question answering on BELEBELE \citep{belebele}, closed-book question answering on EXAMS \citep{exams}, and a variety of natural language understanding benchmarks (XNLI \citep{xnli}, XStoryCloze \citep{lin2022fewshot}, XCOPA \citep{xcopa}, XWinograd \citep{xwino}, and PAWS-X \citep{xpaws}).

All quantitative evaluations are performed on our adapted models after continuous pretraining, but before the alignment stage. We evaluate each checkpoint only on the language that it was trained on. Note that not all of our target languages are covered across all benchmarks. However, each language we examine has evaluations in at least 4 of these benchmarks. We ensured that perplexity measurements were done on a held out set in the target language, and verify that evaluating perplexity on different domains of text such as Wikipedia and MC4 \citep{mc4} have very similar results in appendix \ref{MRT}.

\subsubsection{Quantitative Results}
\begin{figure}[h]
\centering
\includegraphics[width=14cm]{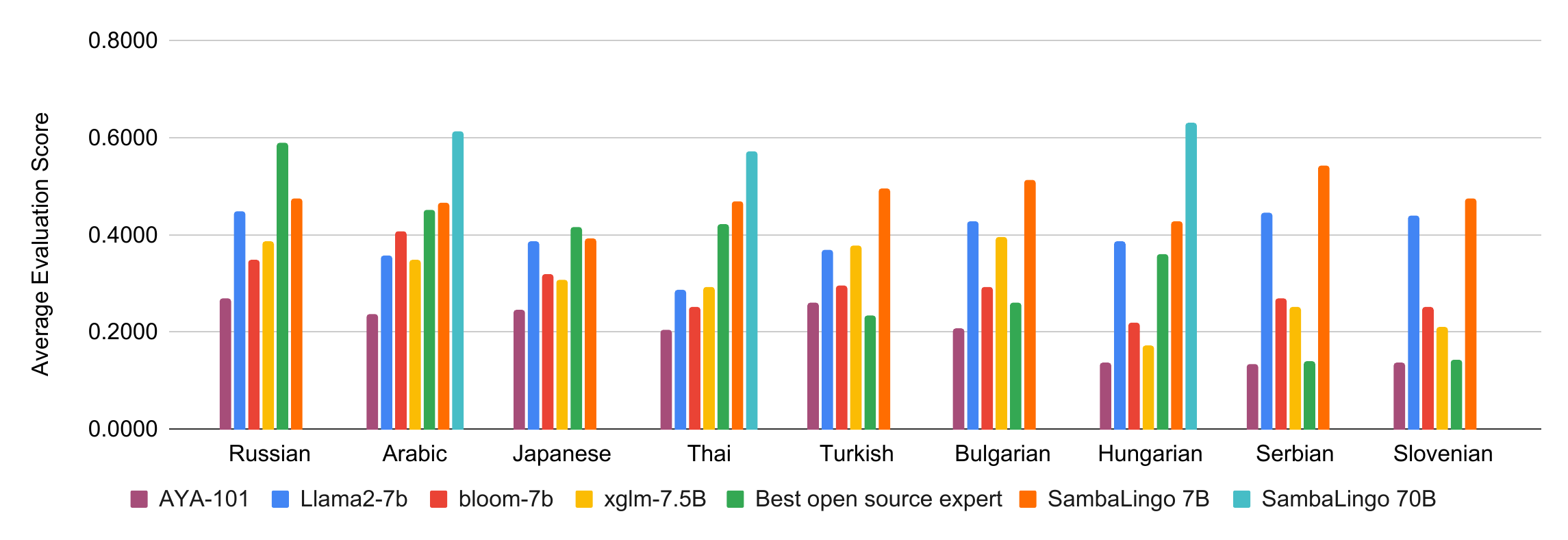}
    \caption{Quantitative evaluation results. The ``best open source experts'' are the same as ones specified in Figure \ref{fig:perplexity_results}. See Appendix \ref{MRT} for the full breakdown.}
    \label{fig:Main_Results}
\end{figure}
We compare our continuously pretrained models against the best open source models available in each target language and state of the art multilingual models. Figure \ref{fig:perplexity_results} shows that our SambaLingo models have a lower perplexity across all existing baselines on a holdout set from our training data. Perplexity on other domains also follows the same trend as shown in appendix \ref{MRT}. Figure \ref{fig:Main_Results} shows the average evaluation score across the evaluation benchmarks introduced in Section \ref{Quantitative Evaluation}, where we see our models outperform all other models in 7/9 languages.

\subsection{Scaling to 70B}\label{Scaling From 7B to 70B}
\begin{table}[h]
\centering
\resizebox{\textwidth}{!}{%
\begin{tabular}{llrrrrrll}
\hline
\rowcolor[HTML]{C0C0C0} 
\multicolumn{1}{c}{\cellcolor[HTML]{C0C0C0}\textbf{Language}} & \multicolumn{1}{c}{\cellcolor[HTML]{C0C0C0}\textbf{Checkpoint}} & \multicolumn{1}{c}{\cellcolor[HTML]{C0C0C0}\textbf{ppl} $(\downarrow)$} & \multicolumn{1}{c}{\cellcolor[HTML]{C0C0C0}\textbf{FLORES EN$\rightarrow$X} $(\uparrow)$} & \multicolumn{1}{c}{\cellcolor[HTML]{C0C0C0}\textbf{FLORES X$\rightarrow$EN} $(\uparrow)$} & \multicolumn{1}{c}{\cellcolor[HTML]{C0C0C0}\textbf{Belebele} $(\uparrow)$} & \multicolumn{1}{c}{\cellcolor[HTML]{C0C0C0}\textbf{SIB-200} $(\uparrow)$} & \multicolumn{1}{c}{\cellcolor[HTML]{C0C0C0}\textbf{XNLI} $(\uparrow)$} & \multicolumn{1}{c}{\cellcolor[HTML]{C0C0C0}\textbf{XStoryCloze} $(\uparrow)$} \\
\hline
\rowcolor[HTML]{EFEFEF} 
\multicolumn{1}{l}{\cellcolor[HTML]{EFEFEF}Arabic} & 70B & \textbf{1.44} & \textbf{54.25} & \textbf{65.60} & \textbf{0.78} & \textbf{0.69} & 0.33 & \textbf{0.68} \\
\rowcolor[HTML]{EFEFEF} 
\multicolumn{1}{l}{\cellcolor[HTML]{EFEFEF}} & 7B & \textbf{1.44} & 53.67 & 61.66 & 0.29 & 0.26 & \textbf{0.34} & 0.65 \\
\rowcolor[HTML]{FFFFFF} 
\multicolumn{1}{l}{\cellcolor[HTML]{FFFFFF}Hungarian} & 70B & \textbf{1.57} & \textbf{58.81} & \textbf{64.03} & \textbf{0.82} & \textbf{0.64} & - & - \\
\rowcolor[HTML]{FFFFFF} 
\multicolumn{1}{l}{\cellcolor[HTML]{FFFFFF}} & 7B & 1.63 & 52.70 & 58.31 & 0.33 & 0.25 & - & - \\
\hline
\end{tabular}%
}
\caption{This table compares compute matched 7B and 70B checkpoints. We look at intermediate checkpoint results and compare 7B models trained for 40B tokens with 70B models trained for 4B tokens.}
\label{Compute Matched: scaling from 7B to 70B}
\end{table}

Scaling to 70B consistently leads to better results as seen in table \ref{fig:Main_Results}. The 70B models in the table have trained on fewer tokens than the 7B models. 

Additionally, we evaluate compute-matched checkpoints of our 7B and 70B Llama 2 models in table \ref{Compute Matched: scaling from 7B to 70B}. The compute-matched 70B checkpoints are trained for 10x fewer steps (4B tokens vs 40B tokens) and perform as well as or better than 7B variants trained over 40B tokens in every benchmark across Arabic and Hungarian. 

\subsection{Evaluating Human Aligned Checkpoints}\label{EvaluatingHACheckpoints}
\subsubsection{GPT-4 as a Judge}
To test our human aligned models' ability to generate high quality responses to real user prompts, we use GPT-4 \citep{openai2024gpt4} as a judge. This method was first introduced by \citet{zheng2023judging} to evaluate English models, and then used by \citet{ustun2024aya} as a method to evaluate multilingual models. The resulting model generations are shuffled and fit to the prompting style suggested by \citep{zheng2023judging} before being fed to GPT-4. See Appendix \ref{app:qual}  for the manually collected prompts and section \ref{Qualitative Evaluation} for the evaluation results.

GPT-4 as a judge has been widely accepted by the community as a way to evaluate chat models \citep{zheng2023judging, verga2024replacing}, and we extend this to multilingual models. To ensure that GPT-4 is understanding the multilingual text we have native speakers read through a few examples of GPT-4 explaining its decision making process. The native speakers unanimously agree that GPT-4 clearly understands the content in other languages. In appendix \ref{Appendix: GPT-4 As A Judge} we include example model generations along with GPT-4's corresponding preferences and explanations. Further work is needed to do a large scale study to see how GPT-4 preferences align with human preferences in other languages. 

\subsubsection{Qualitative Results}\label{Qualitative Evaluation}
\begin{figure}[h]
\centering
    \begin{subfigure}{0.4\textwidth}
        \includegraphics[width=\textwidth]{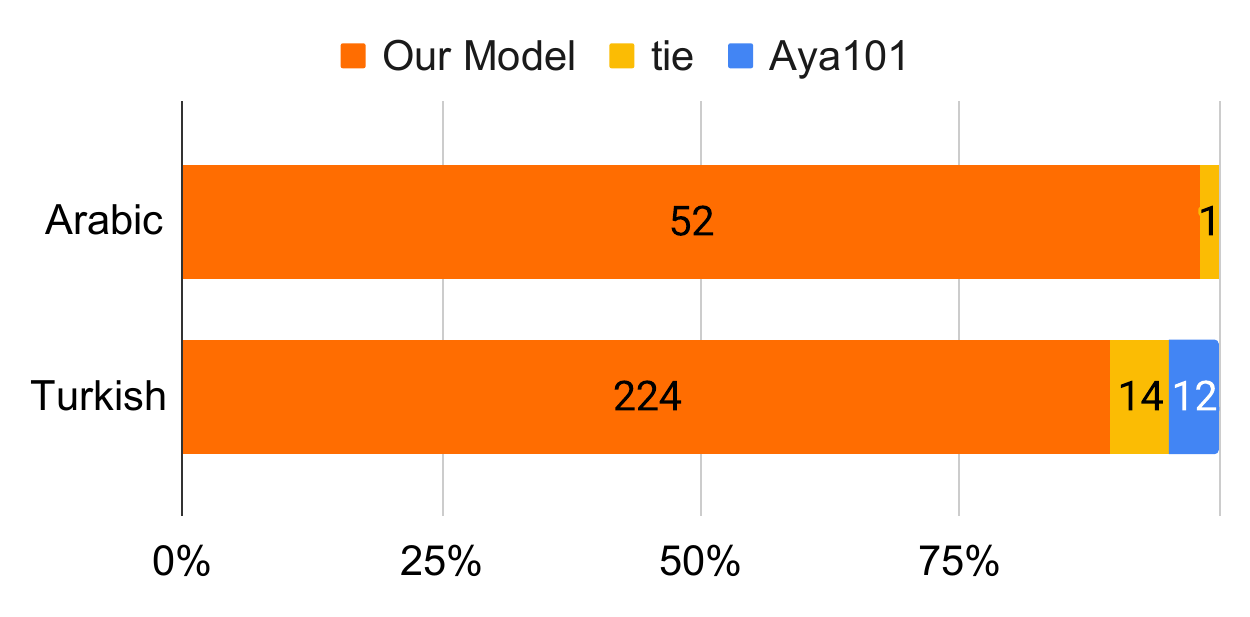}
        \caption{SambaLingo vs Aya101}
    \end{subfigure}%
    \begin{subfigure}{0.4\textwidth}
        \includegraphics[width=\textwidth]{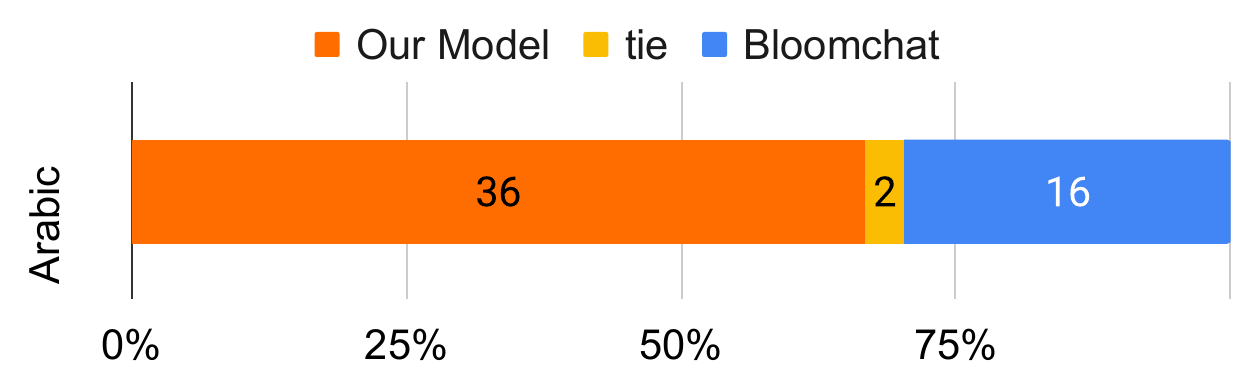}
        \caption{SambaLingo vs BloomChat-v1}
    \end{subfigure}
     \begin{subfigure}{0.4\textwidth}
        \includegraphics[width=\textwidth]{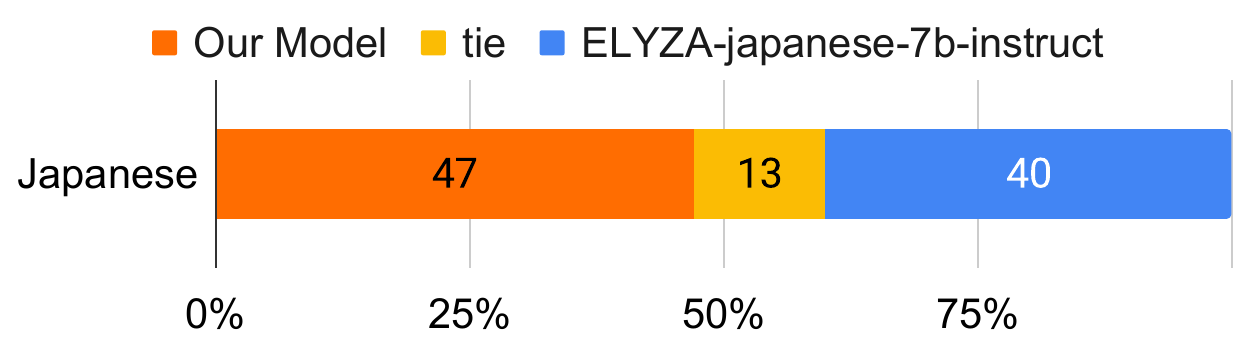}
        \caption{SambaLingo vs ELEYZA-7b-instruct}
    \end{subfigure}%
    \begin{subfigure}{0.4\textwidth}
        \includegraphics[width=\textwidth]{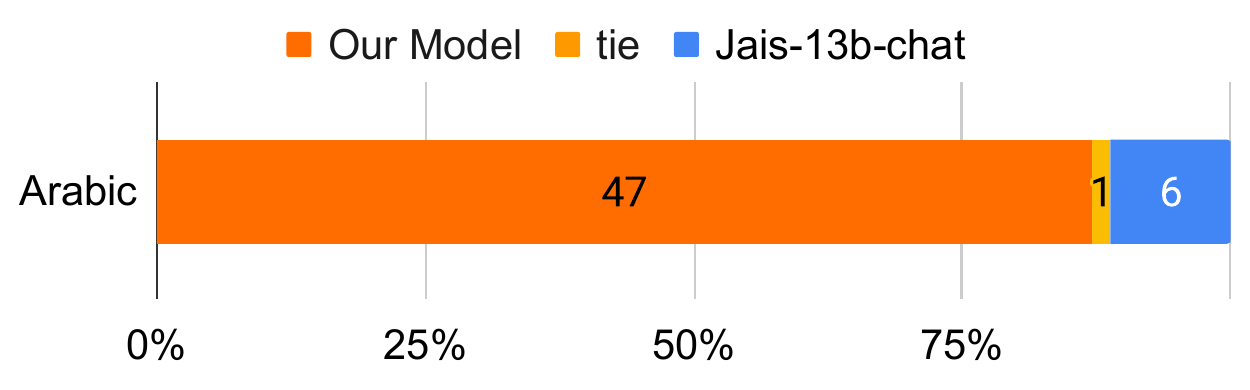}
        \caption{SambaLingo vs Jais-13b-chat}
    \end{subfigure}
    \caption{GPT4 evaluation result}
    \label{fig:human-eval}
\end{figure}
 Measuring win-rate using GPT-4 as a judge only works in scenarios where a human aligned or instruction tuned model is available in a language. Given this constraint, we were only able to find relevant comparisons for Arabic, Japanese and Turkish, and do not have qualitative evaluations for our models in the other 6 languages. We do not compare to llama2-chat because we found that Llama2-chat and other open source English foundation chat models reply in English when prompted in the target language, instead of replying back in the target language. The results of our evaluation are shown in Figure \ref{fig:human-eval}. Our SambaLingo models consistently out-perform other models in the same language. For details about the native speaker-curated prompts, see Appendix \ref{app:qual}. We additionally run evaluations with Claude Opus \citep{anthropic2024claude} as a judge to ensure that there is no bias by GPT-4 and find very similar results in appendix \ref{app:claude}

\section{Ablations}
\label{Ablations}
In this section, we present ablations of our design decisions in Section \ref{Adaptation Methodology}. Section \ref{VocabularyExpansion} presents experiments motivating the modifications we make to the base model's tokenizer and how we initialize its new embeddings. Section \ref{Direct Preference Optimization} ablates the amount of target language data and use of machine translated data in the DPO phase of our methodology. Finally, section \ref{Importance Of Base Model Quality} looks at the impact of the quality of the base model.
\subsection{Vocabulary Expansion}\label{VocabularyExpansion}
The Llama2 tokenizer is centered towards English. While this tokenizer can encode characters in any language, it will be very inefficient for non-English text.  In fact, the BPE tokenizer may tokenize non-Latin characters as multiple independent bytes. One way to mitigate this problem is to extend the vocabulary of the base model by adding new tokens that represent the target language to it, and start adaptation training with this expanded vocabulary. This method also helps improve the inference efficiency in the target language. We explore different sizes for the expanded vocabulary and their impacts on fertility \cite{acs2019} in Table \ref{Added Tokens vs Language PPL Table} and Figure \ref{fig:token_fertility}. We chose to expand the vocabulary by 25,000 tokens for all languages as it yields the lowest fertility for all languages and highest throughput on the hardware platform. 
\begin{table}[h]
\centering
\resizebox{\textwidth}{!}{%
\begin{tabular}{llrrrrrll}
\hline
\rowcolor[HTML]{C0C0C0} 
\multicolumn{1}{c}{\cellcolor[HTML]{C0C0C0}\textbf{Added Tokens}} & \multicolumn{1}{c}{\cellcolor[HTML]{C0C0C0}\textbf{Hungarian}} & \multicolumn{1}{c}{\cellcolor[HTML]{C0C0C0}\textbf{Russian}} & \multicolumn{1}{c}{\cellcolor[HTML]{C0C0C0}\textbf{Turkish}} & \multicolumn{1}{c}{\cellcolor[HTML]{C0C0C0}\textbf{Bulgarian}} & \multicolumn{1}{c}{\cellcolor[HTML]{C0C0C0}\textbf{Arabic}} & \multicolumn{1}{c}{\cellcolor[HTML]{C0C0C0}\textbf{Japanese}} & \multicolumn{1}{c}{\cellcolor[HTML]{C0C0C0}\textbf{Thai}} \\
\hline
\rowcolor[HTML]{EFEFEF} 
\multicolumn{1}{l}{\cellcolor[HTML]{EFEFEF}0} & 2.70 & 2.28 & 3.28 & 2.36 & 4.23 & 2.07 & 4.84 \\
\rowcolor[HTML]{FFFFFF} 
\multicolumn{1}{l}{\cellcolor[HTML]{FFFFFF}1000} & 2.52 & 2.25 & 2.56 & 2.19 & 2.11 & 1.75 & 2.10 \\
\rowcolor[HTML]{EFEFEF} 
\multicolumn{1}{l}{\cellcolor[HTML]{EFEFEF}4000} & 2.14 & 2.05 & 2.20 & 1.92 & 1.67 & 1.23 & 1.50 \\
\rowcolor[HTML]{FFFFFF} 
\multicolumn{1}{l}{\cellcolor[HTML]{FFFFFF}25000} & 1.78 & 1.78 & 1.77 & 1.66 & 1.26 & 0.93 & 1.10 \\
\hline
\end{tabular}%
}
\caption{Number of added tokens vs fertility (average number of tokens per "word")}
\label{Added Tokens vs Language PPL Table}
\end{table}

\subsubsection{Vocabulary Expansion vs Original Tokenizer}\label{Vocabulary Expansion vs Original Tokenizer}
To measure the impact of vocabulary expansion on accuracy, we train two models—one using an expanded vocabulary and the other using the original vocabulary—across two three languages: Hungarian, Arabic and Serbian. We find that expanding the vocabulary does not have significant impact on the downstream accuracy. Nonetheless, given the benefit that the expanded vocabulary has for inference and sequence length utilization in the target language, we chose to expand the vocabulary of the base model.

\begin{table}[h]
\centering
\resizebox{\textwidth}{!}{%
\begin{tabular}{llrrrrrll}
\hline
\rowcolor[HTML]{C0C0C0} 
\multicolumn{1}{c}{\cellcolor[HTML]{C0C0C0}\textbf{Language}} & \multicolumn{1}{c}{\cellcolor[HTML]{C0C0C0}\textbf{Tokenizer}} & \multicolumn{1}{c}{\cellcolor[HTML]{C0C0C0}\textbf{ppl $(\downarrow)$}} & \multicolumn{1}{c}{\cellcolor[HTML]{C0C0C0}\textbf{FLORES EN$\rightarrow$X} $(\uparrow)$} & \multicolumn{1}{c}{\cellcolor[HTML]{C0C0C0}\textbf{FLORES X$\rightarrow$EN} $(\uparrow)$} & \multicolumn{1}{c}{\cellcolor[HTML]{C0C0C0}\textbf{Belebele} $(\uparrow)$} & \multicolumn{1}{c}{\cellcolor[HTML]{C0C0C0}\textbf{SIB-200} $(\uparrow)$} & \multicolumn{1}{c}{\cellcolor[HTML]{C0C0C0}\textbf{XNLI} $(\uparrow)$} & \multicolumn{1}{c}{\cellcolor[HTML]{C0C0C0}\textbf{XStoryCloze} $(\uparrow)$} \\
\hline
\rowcolor[HTML]{EFEFEF} 
\multicolumn{1}{l}{\cellcolor[HTML]{EFEFEF}Arabic} & Original & 1.50 & 48.27 & 57.35 & 0.27 & 0.27 & \textbf{0.34} & 0.63 \\ 
\rowcolor[HTML]{EFEFEF} \multicolumn{1}{l}{\cellcolor[HTML]{EFEFEF}} & Expanded & \textbf{1.46} & \textbf{52.66} & \textbf{61.05} & \textbf{0.32} & \textbf{0.35} & \textbf{0.34} & \textbf{0.64} \\
\rowcolor[HTML]{FFFFFF} 
\multicolumn{1}{l}{\cellcolor[HTML]{FFFFFF}Hungarian} & Original & \textbf{1.61} & \textbf{52.70} & \textbf{58.31} & \textbf{0.33} & 0.26 & - & - \\
\rowcolor[HTML]{FFFFFF} 
\multicolumn{1}{l}{\cellcolor[HTML]{FFFFFF}} & Expanded & 1.63 & 51.82 & 57.12 & 0.30 & \textbf{0.34} & - & - \\
\rowcolor[HTML]{EFEFEF} 
\multicolumn{1}{l}{\cellcolor[HTML]{EFEFEF}Serbian} & Original & \textbf{1.403} & 56.15 & 64.89 & 0.32 & \textbf{0.59} & - & - \\ 
\rowcolor[HTML]{EFEFEF} \multicolumn{1}{l}{\cellcolor[HTML]{EFEFEF}} & Expanded & 1.435  & \textbf{58.30} & \textbf{66.3}5 & \textbf{0.37} & 0.52 & - & - \\
\hline
\end{tabular}%
}
\caption{Accuracy after training with expanded vocabulary vs original tokenizer}
\label{Vocabulary Expansion Ablation Table}
\end{table}

\subsubsection{Initializing new token embeddings}\label{New Token Embedding Initialization}

We experiment with 4 different token initialization strategies for the new tokens added to the vocabulary across 3 languages - Hungarian Arabic and Thai. For each experiment, we train the model for 10 billion tokens and compare the loss values. Let $V$ be the set of tokens in the original vocabulary, and $E(t)$ the embedding vector of a token $t\in V$. The four token initialization methods we consider are as follows:

\begin{itemize}
    \item \texttt{gaussian}: $\mathcal{N}(0, 0.02)$
    \item \texttt{xavier\_uniform}: a uniform initialization $\mathcal{U}(-\alpha, \alpha)$, where $\alpha = \sqrt{\frac{6}{\textrm{fan\_in}+\textrm{fan\_out}}}$. Introduced by \cite{glorot2010understanding} and used by \cite{csaki2023efficiently} in Hungarian and Thai language adaptation
    \item \texttt{avg\_all} \cite{hewitt2021initializing}: For each new token $t'$, initialize $E(t') = \textrm{mean}(\{E(t) \forall t \in V\})$ 
    \item \texttt{avg\_subwords} \citep{liu2024chipnemo, koto2021indobertweet}: For each new token $t'$, let $L_{t'} = [t_1,...,t_k]$ be the list of $k$ tokens that $t'$ would have been tokenized as under the original tokenizer. Initialize the new embedding with $E(t') = \textrm{mean}([E(t_1),...,E(t_k)])$.
\end{itemize}

\begin{figure*}[h]
    \centering
    \begin{subfigure}[h]{0.5\textwidth}
        \includegraphics[width=\linewidth]{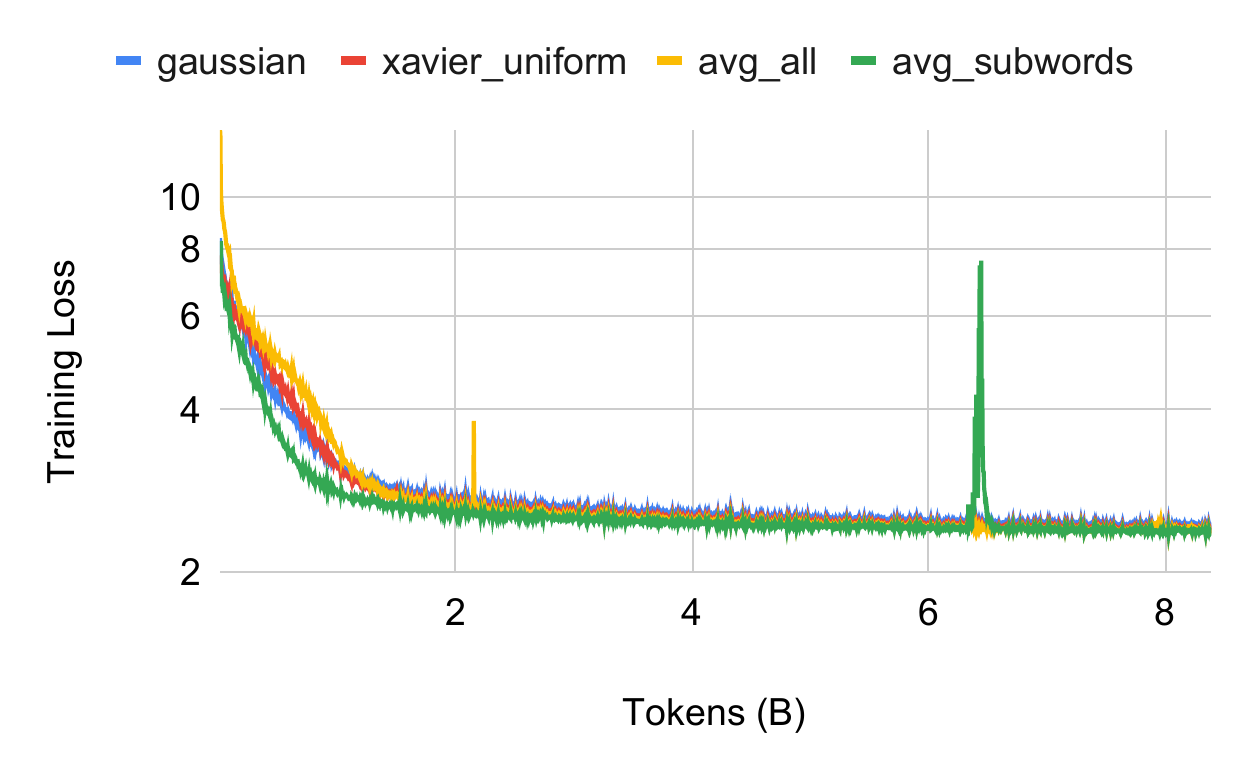}
        \subcaption[]{Arabic}
    \end{subfigure}%
    ~
    \begin{subfigure}[h]{0.5\textwidth}
        \includegraphics[width=\linewidth]{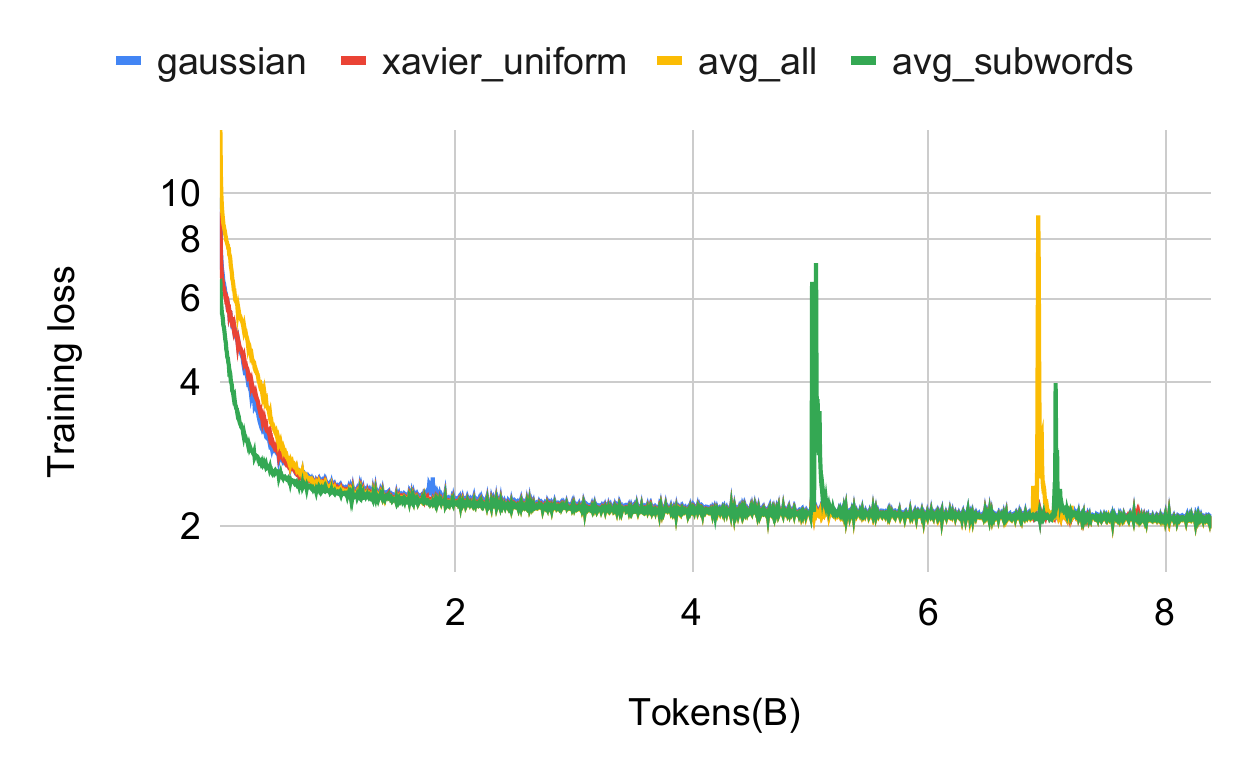}
        \subcaption[]{Hungarian}
    \end{subfigure}
    \caption{Training loss for different token initialization methods}
    \label{fig:tokinit_trainloss}
\end{figure*}

\begin{table*}[h]
\centering
\resizebox{\textwidth}{!}{%
\begin{tabular}{llrrrrrll}
\hline
\rowcolor[HTML]{C0C0C0} 
\multicolumn{1}{c}{\cellcolor[HTML]{C0C0C0}\textbf{Language}} & \multicolumn{1}{c}{\cellcolor[HTML]{C0C0C0}\textbf{Initialization Method}} & \multicolumn{1}{c}{\cellcolor[HTML]{C0C0C0}\textbf{ppl} $(\downarrow)$} & \multicolumn{1}{c}{\cellcolor[HTML]{C0C0C0}\textbf{FLORES EN$\rightarrow$X} $(\uparrow)$} & \multicolumn{1}{c}
{\cellcolor[HTML]{C0C0C0}\textbf{FLORES X$\rightarrow$EN} $(\uparrow)$} & \multicolumn{1}{c}{\cellcolor[HTML]{C0C0C0}\textbf{Belebele} $(\uparrow)$} & \multicolumn{1}{c}{\cellcolor[HTML]{C0C0C0}\textbf{SIB-200} $(\uparrow)$} & \multicolumn{1}{c}{\cellcolor[HTML]{C0C0C0}\textbf{XNLI} $(\uparrow)$} & \multicolumn{1}{c}{\cellcolor[HTML]{C0C0C0}\textbf{XStoryCloze} $(\uparrow)$} \\
\hline
\rowcolor[HTML]{EFEFEF} 
\multicolumn{1}{l}{\cellcolor[HTML]{EFEFEF}Arabic} & \texttt{gaussian} & 1.50 & 48.48 & 57.31 & 0.34 & 0.25 & 0.34 & 0.61 \\
\rowcolor[HTML]{EFEFEF} \multicolumn{1}{l}{\cellcolor[HTML]{EFEFEF}} & \texttt{xavier\_uniform} & 1.49 & 50.46 & 58.90 & 0.36 & 0.26 &  0.33 & 0.62 \\
\rowcolor[HTML]{EFEFEF} \multicolumn{1}{l}{\cellcolor[HTML]{EFEFEF}} & \texttt{avg\_all} & \textbf{1.48} & 50.54 & 58.29 & 0.34 & 0.25 & \textbf{0.35} & 0.63 \\
\rowcolor[HTML]{EFEFEF} \multicolumn{1}{l}{\cellcolor[HTML]{EFEFEF}} & \texttt{avg\_subwords} & \textbf{1.48} & \textbf{50.87} & \textbf{59.62} & \textbf{0.38} & \textbf{0.27} & 0.34 & \textbf{0.64} \\
\rowcolor[HTML]{FFFFFF} 
\multicolumn{1}{l}{\cellcolor[HTML]{FFFFFF}Hungarian} & \texttt{gaussian} & \textbf{1.65} & \textbf{51.42} & \textbf{56.92} & 0.32 & \textbf{0.50} & - & - \\
& \texttt{xavier\_uniform} & \textbf{1.65} & 49.52 & 55.81 & \textbf{0.34} & 0.42 & - & - \\
& \texttt{avg\_all} & 1.76 & 51.39 & 56.86 & \textbf{0.34} & 0.45 & - & - \\
& \texttt{avg\_subwords} & \textbf{1.65} & 50.79 & 56.77 & 0.33 & 0.30 & - & - \\
\rowcolor[HTML]{EFEFEF} 
\multicolumn{1}{l}{\cellcolor[HTML]{EFEFEF}Thai} & \texttt{gaussian} & 1.31 & 51.50 & 52.95 & 0.33 & 0.53 & 0.44 & - \\
\rowcolor[HTML]{EFEFEF} \multicolumn{1}{l}{\cellcolor[HTML]{EFEFEF}} & \texttt{xavier\_uniform} & 1.31 & 52.88 & 55.34 & 0.32 & 0.30 &  0.38 & - \\
\rowcolor[HTML]{EFEFEF} \multicolumn{1}{l}{\cellcolor[HTML]{EFEFEF}} & \texttt{avg\_all} & 1.31 & 52.89 & 55.36 & 0.35 & \textbf{0.60} & \textbf{0.46} & - \\
\rowcolor[HTML]{EFEFEF} \multicolumn{1}{l}{\cellcolor[HTML]{EFEFEF}} & \texttt{avg\_subwords} & \textbf{1.30} & \textbf{53.34} & \textbf{55.36} & \textbf{0.37} & 0.35 & \textbf{0.46} & - \\
\hline

\end{tabular}%
}
\caption{Multilingual evaluations across token embedding initialization methods}
\label{tab:tokinit_eval}
\end{table*}

Figure \ref{fig:tokinit_trainloss} shows that after continuous pretraining for 10B tokens, all methods converge to similar loss values, with \texttt{avg\_subwords} showing faster convergence. Table \ref{tab:tokinit_eval} shows the impact on downstream benchmarks. For Thai and Arabic, \texttt{avg\_subwords} achieves marginally better scores while for Hungarian the results are mixed. These results show that the choice of initialization has minimal impact on the accuracy of end model when trained for 10 billion tokens. However \texttt{avg\_subwords} gives faster training loss convergence, so we chose to initialize the new embeddings using \texttt{avg\_subwords}.

\subsection{Direct Preference Optimization}\label{Direct Preference Optimization}

\subsubsection{DPO Data Mixture}\label{DPO Data Mixture}

There is a lack of supervised finetuning and human alignment data across different languages. Collecting such data can be difficult and expensive. Given that the models obtained from our methodology are bilingual, we explore the question of how much of the human alignment data can be English and how much of it has to be from the target language. We run DPO on data mixtures of the English/Target language data ratio across 100:1, 10:1, 10:3 and 1:1, and observe the resulting win-rate in pairwise comparisons with the model trained on a 1:1 data ratio. For each experiment we keep the amount of English data the same and downscale the target language. We run these experiments on two languages: Hungarian and Arabic, with results in Table \ref{tab:DPO data}. We show that a 10:1 data ratio can perform almost as well as 1:1 data ratio for Hungarian. For Arabic, even a 10:3 data ratio still falls behind the performance of 1:1. One hypothesis is that Hungarian is more linguistically similar to English than Hungarian so there is more language transfer during fine tuning, but further research is needed to understand how the language impacts optimal alignment data mixture ratio. 

\begin{table*}[h]
\begin{center}
\begin{tabular}{|l|l|l|l|l|}
\hline
Target Language: English Ratio  & 100:1 & 10:1 & 10:3 & 1:1 \\
\hline
Arabic        & 30.39\% & 35.00\% & 34.62\% & 50.00\% \\
\hline
Hungarian     & 39.29\% & 45.18\% & 45.78\% & 50.00\% \\
\hline
\end{tabular}
\end{center}
\caption{DPO data mixture result (win-rate compared with 1:1 data mixture)}
\label{tab:DPO data}
\end{table*}
\subsubsection{Impact of Translated Human Preference Data}\label{abl:dpo}

Results in Table \ref{tab:DPO data} are based on translated data from the target language. \citet{ustun2024aya} emphasized the importance of human written prompt completion pairs and claim that translated data does not perform as well. However, their work does not start with a high quality pretrained base model, nor do they use DPO. In order to understand whether machine translated data is a viable option for human alignment, we explore the impact of alignment using both approaches. We use Google translated \texttt{ultrafeedback-200k} data for one run and human-written data from Open Assistant Conversations (OASST1) \citep{köpf2023openassistant} for the other. We run this study on Russian, as it is has the most human written data from OASST1 \citep{köpf2023openassistant}. The model trained using translated data attains a 50.47\% win rate compared to the model trained with OASST1. This comparison does not control for the diversity and quality of the question answer pairs in the dataset because chat datasets with parallel human translated data in multiple languages. so this comparison is not meant to illustrate that translated data is as good or better than native data, but rather to show that human written data is not a silver bullet required to obtain good quality aligned models in other languages.

\subsection{Importance Of Base Model Quality}\label{Importance Of Base Model Quality}
To explore the relationship between the quality of the base model employed for language adaptation and its subsequent impact on accuracy in the target language, we ablate using two different base models - Llama 2 7B and GPT-13B \citep{srinivasan2023training}. The GPT-13B model is trained on much fewer tokens compared to llama2. We measure the GPT-13B model on some commonly accepted English benchmarks instead of our multilingual evaluation suite because these benchmarks are used more frequently to compare English checkpoints. GPT-13B lags behind Llama 2 7B in every English evaluation tasks we measured in Table \ref{Performance Metrics Table 2}. 

\begin{table*}[h]
\centering
\resizebox{\textwidth}{!}{%
\begin{tabular}{llrrrrrll}
\hline
\rowcolor[HTML]{C0C0C0} 
\multicolumn{1}{c}{\cellcolor[HTML]{C0C0C0}\textbf{Base Model}} & \multicolumn{1}{c}{\cellcolor[HTML]{C0C0C0}\textbf{ppl}$(\downarrow)$} & \multicolumn{1}{c}{\cellcolor[HTML]{C0C0C0}\textbf{FLORES EN$\rightarrow$X}$(\uparrow)$} & \multicolumn{1}{c}{\cellcolor[HTML]{C0C0C0}\textbf{FLORES X$\rightarrow$en$(\uparrow)$}} & \multicolumn{1}{c}{\cellcolor[HTML]{C0C0C0}\textbf{Belebele}$(\uparrow)$} & \multicolumn{1}{c}{\cellcolor[HTML]{C0C0C0}\textbf{SIB-200}$(\uparrow)$} \\
\hline
\rowcolor[HTML]{EFEFEF} 
\multicolumn{1}{l}{\cellcolor[HTML]{EFEFEF}GPT-13B} & 1.80 & 37.94 & 48.99 & 0.28 & \textbf{0.25} \\ 
\rowcolor[HTML]{EFEFEF} 
\multicolumn{1}{l}{\cellcolor[HTML]{EFEFEF}Llama-2-7b} & \textbf{1.61} & \textbf{53.72} & \textbf{58.65} & \textbf{0.34} & \textbf{0.25} \\
\hline
\end{tabular}%
}
\caption{Performance of GPT-13B and Llama 2 7B on Hungarian benchmarks after adaptation}
\label{Performance Metrics Table}
\end{table*}

We adapt both of these models to Hungarian. Table \ref{Performance Metrics Table} illustrates that using a higher quality base model (Llama 2 7B) leads to better downstream performance in the target language. These results show that many of the benefits of training come from the base model quality not just the continuous training we do. This additionally indicates that as newer higher quality models become available, there is value in applying our proposed adaptation methodology on new base models.

\section{Limitations}
\label{limitations}
Our work has several limitations, including the need for extensive data from the target language, which is often unavailable for many languages. We study 9 diverse languages, but further research is required to address multilingual data scarcity. Due to compute and time constraints, our ablation studies focus on around 3 languages each, assuming similar results for other languages, although linguistic diversity and data availability may affect this. Additionally, we evaluate our chat-based model using GPT-4 as a judge, and while this has been shown to strongly correlate with human preferences in English, we are uncertain how well this works in other languages. We acknowledge that publicly releasing LLMs is risky because they can inadvertently generate harmful or biased content, compromise privacy, and be exploited for malicious purposes such as spreading misinformation. Moreover, while our models are adapted to other languages and cultures, the English base model, data biases, and use of translation may prevent them from fully capturing the nuances of cultures and languages from around the world.

\section{Conclusion}
\label{Conclusion}
We present a methodology to adapt pretrained LLMs to new languages. The methodology encompasses both continuous pretraining and alignment to human preferences in the target language. We present experimental results to justify our design choices and scale our methodology to 9 typologically diverse languages and 2 parameter scales. We make our evaluation scripts and final checkpoints publically available to facilitate future research, and we hope this work outlines a clearer path towards attaining state of the art language models in every language.
\clearpage
\bibliography{colm2024_conference}
\bibliographystyle{colm2024_conference}
\clearpage

\appendix
\section{Hyperparameters}\label{app:hyper}
\begin{itemize}
    \item \textbf{Continuous Pre-training}: We pack the pretraining mixture into sequences of length 4096 and pretrain with \textit{document attention} as described in Section 3.2 of \citet{iyer2022opt} to ensure we only attend to tokens in the context of the corresponding text document. We train with a global batch size of 1024, sequence length of 4096, maximum learning rate of 1e-4 with cosine decay, warm-up ratio of 0.01 and a weight decay of 0.1. Each expert is trained for a maximum of 4 epochs, following \citep{muennighoff2023scaling}. Notably, we train all model parameters, foregoing use of PEFT methods such as LoRA \citep{hu2022lora}, which are known to be inferior to full parameter training \citep{zhao2024galore}\citep{sun2023comparative}.
    \item \textbf{Supervised Finetuning}: We use a global batch size of 512 and a maximum sequence length of 2048 tokens. We used a linear decay learning rate of 2e-5 with 10\% warm up
    \item \textbf{Direct Preference Optimization}: We train with a global batch size 32 for 3 epochs, a linear decay learning rate of 5e-7, 10\% warmup and $\beta= 0.1$  as the regularization
factor for DPO
\end{itemize}

\section{Language Experts vs Monolith Multilingual Model}\label{Monolith Multilingual Model vs Language Experts}

“The Curse Of Multilinguality” \citep{chang2023multilinguality, conneau2020unsupervised} is the idea that LLMs have a fixed capacity with which to learn various languages. This theory claims that as one expands the number of languages a model is trained on, the various languages compete for the capacity of the model, therefore degrading the models performance across all languages. \citet{blevins2024breaking} attempt to address this phenomenon by adapting multiple small-scale language experts from XGLM-1.7B \citep{lin2022fewshot}, one for each language, and show that each expert outperforms training a single monolithic model trained simultaneously on one language. We build on these results by scaling this study to 7B parameters and use more comprehensive evaluation metrics than just perplexity. We compare our 9 Llama 2 7B language experts against a monolith Llama 2 7B model continuously pretrained on all 9 languages. We ensure that each language is represented equally in the monolith's training data and the vocabulary is expanded to represent all 9 languages evenly.

For comparison's sake, we select intermediate model checkpoints such that each individual language expert has used the same amount of compute as the monolith multilingual model. This means that the experts required 9x more compute to train then the monolith. Table \ref{tab:Monolith multilingual continuous training vs language experts} averages the evaluation results across all 9 languages and finds that the monolith model and language experts have very similar performance. This implies that if one wants to adapt to many languages at once, it may be more compute-efficient to continuously train a multi-linugal model rather then independent experts. Further work is warranted to determine how this result scales with an increasing number of target languages.

\begin{table}[b]
\small 
\centering
\begin{tabular}{lccc}
\toprule
\textbf{Benchmark}& \textbf{Llama2-7b} & \textbf{Multilingual} & \textbf{Language} \\
\textbf{(Num Shots)} & \textbf{Avg} & \textbf{Monolith Avg} & \textbf{Expert Avg} \\
\midrule
$\downarrow$ Holdout PPL & 1.75 & 1.55 & \textbf{1.50} \\
$\uparrow$ FLORES X\textgreater{}en(8) & 40.42\% & 50.69\% & \textbf{51.71\%} \\
$\uparrow$ Belebele (3) & \textbf{36.24\%} & 33.36\% & 32.09\% \\
$\uparrow$ SIB-200(3) & 26.67\% & \textbf{38.04\%} & 33.43\% \\
$\uparrow$ XNLI (0) & 39.00\% & \textbf{43.44\%} & 43.04\% \\
$\uparrow$ XStoryCloze (0) & 56.35\% & 65.75\% & \textbf{68.03\%} \\
$\uparrow$ XWinograd (0) & 69.48\% & \textbf{72.39\%} & 71.97\% \\
$\uparrow$ PAWS-X (0) & 51.00\% & \textbf{54.40\%} & 53.50\% \\
$\uparrow$ MGSM (3) & \textbf{5.40\%} & 4.00\% & 4.20\% \\
\bottomrule
\end{tabular}
\caption{Monolith multilingual continuous training vs language experts, averaged over all 9 languages.}
\label{tab:Monolith multilingual continuous training vs language experts}
\end{table}
\clearpage
\section{Expanded Vocabulary Tokenizer Fertility}
\subsection{Expanded Vocabulary Tokenizer Fertility}

\begin{figure}[h]
    \centering
    \includegraphics[width=16cm]{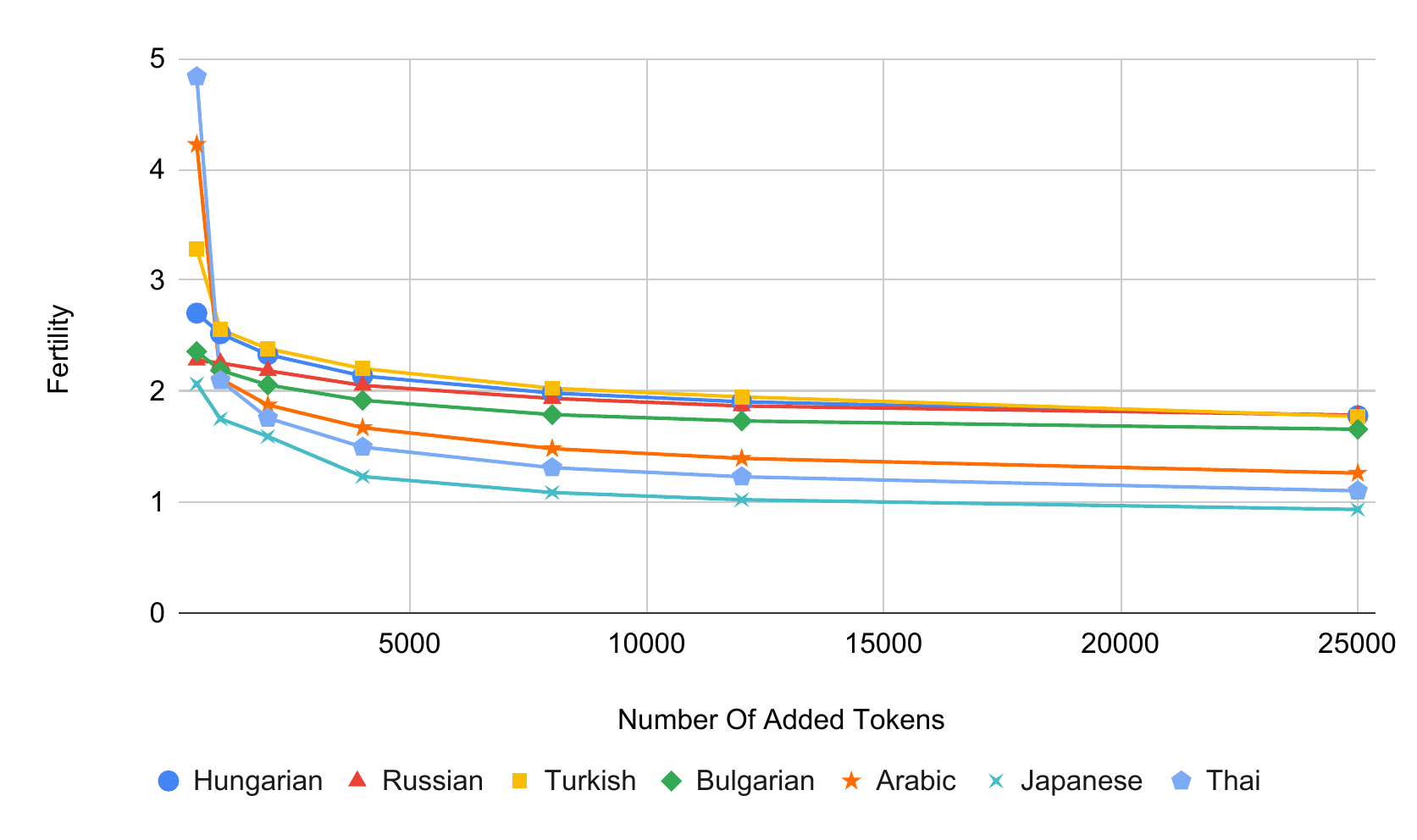}
    \caption{Tokenizer Fertility: the average number of tokens per "word" \citep{acs2019}}
    \label{fig:token_fertility}
\end{figure}

In figure \ref{fig:token_fertility} We measure the fertility of the tokenizer as we expand the vocabulary, and see that we can improve the fertility from about 4.8 to 1.1 on Thai. This is about a 4.35x improvement, implies that inference speeds can improve up to 4.35x compared to the Llama2 tokenizer. 

\section{Base Model English Evaluation}
\label{English Base Model Eval}
\begin{table}[h]
\centering
\resizebox{\textwidth}{!}{%
\begin{tabular}{llrrrrrll}
\hline
\rowcolor[HTML]{C0C0C0} 
\multicolumn{1}{c}{\cellcolor[HTML]{C0C0C0}} & \multicolumn{1}{c}{\cellcolor[HTML]{C0C0C0}\textbf{HellaSwag}$(\uparrow)$} & \multicolumn{1}{c}{\cellcolor[HTML]{C0C0C0}\textbf{OpenBookQA}$(\uparrow)$} & \multicolumn{1}{c}{\cellcolor[HTML]{C0C0C0}\textbf{ARC-E}$(\uparrow)$} & \multicolumn{1}{c}{\cellcolor[HTML]{C0C0C0}\textbf{ARC-C}$(\uparrow)$} & \multicolumn{1}{c}{\cellcolor[HTML]{C0C0C0}\textbf{PiQA}$(\uparrow)$} & \multicolumn{1}{c}{\cellcolor[HTML]{C0C0C0}\textbf{Winogrande}$(\uparrow)$} \\
\hline
\rowcolor[HTML]{EFEFEF} 
\multicolumn{1}{l}{\cellcolor[HTML]{EFEFEF}GPT-13B} & 0.60 & 0.36 & 0.53 & 0.30 & 0.76 & 0.60 \\ 
\rowcolor[HTML]{EFEFEF} 
\multicolumn{1}{l}{\cellcolor[HTML]{EFEFEF}Llama-2-7B} & \textbf{0.76} & \textbf{0.57} & \textbf{0.73} & \textbf{0.48} & \textbf{0.80} & \textbf{0.70} \\
\hline
\end{tabular}%
}
\caption{Performance of GPT-13B and Llama-2-7B on English NLU benchmarks}
\label{Performance Metrics Table 2}
\end{table}
\clearpage
\section{Qualitative Results} \label{app:qual}
For Arabic, we compare our 7B arabic expert with aya-101 \citep{ustun2024aya}, Jais-13b-chat \citep{sengupta2023jais}, and Bloomchat-v1 \citep{bloomchat} and use prompts from x-self-instruct-seed-32 \citep{xinstruct} and xOA22 \citep{xoa}. Our Arabic chat model reaches 87.96\% win rate compared to Jais-13B-chat, 99.06\% win rate compared to Aya101, and 68.52\% compared to Bloomchat-v1. For Japanese, we compare our Japanese chat model with ELYZA-japanese-Llama-2-7b-instruct \citep{elyzallama2023} on 100 randomly sampled prompts aya dataset \citep{ustun2024aya},  reaching a win rate of 53.5\% For Turkish, we compare our Turkish chat model against aya-101 \citep{ustun2024aya} using prompts from the test set of aya dataset \citep{ustun2024aya}, leading to win-rate of 92.4\%.
\subsection{Evaluating Chat Models With Claude} \label{app:claude}
\begin{figure}[h]
\centering
    \begin{subfigure}{0.4\textwidth}
        \includegraphics[width=\textwidth]{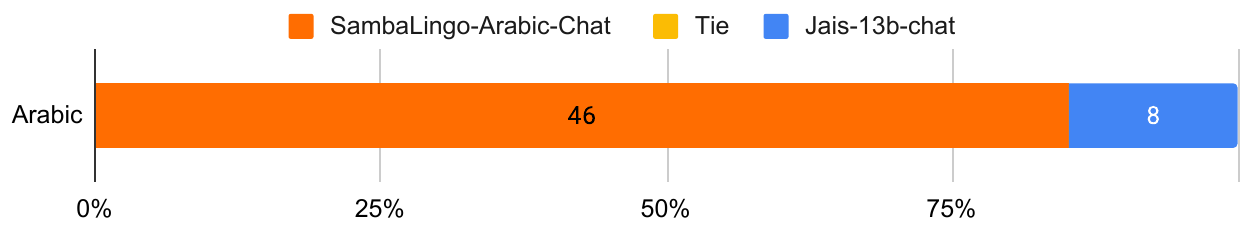}
        \caption{SambaLingo vs Jais-13b-chat}
    \end{subfigure}%
    \begin{subfigure}{0.4\textwidth}
        \includegraphics[width=\textwidth]{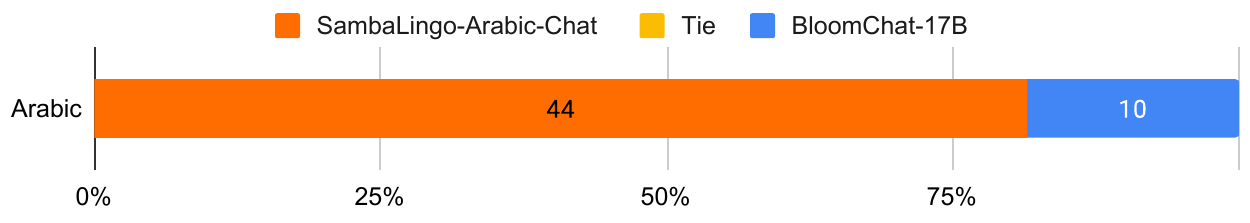}
        \caption{SambaLingo vs BloomChat-v1}
    \end{subfigure}
     \begin{subfigure}{0.4\textwidth}
        \includegraphics[width=\textwidth]{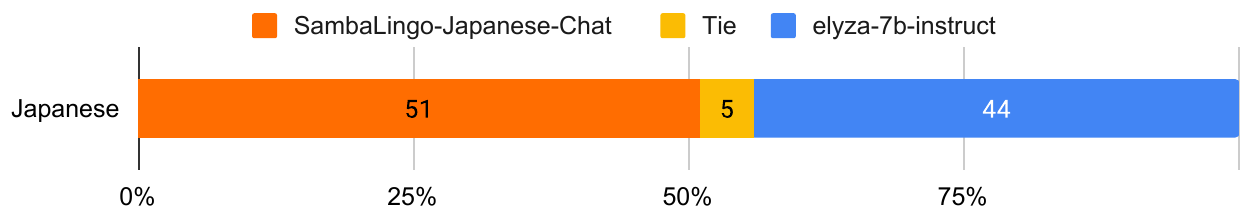}
        \caption{SambaLingo vs ELEYZA-7b-instruct}
    \end{subfigure}%
    \caption{Claude Opus evalution results}
    \label{fig:claude-eval}
\end{figure}

We run evaluations using the same prompt as GPT-4 as a judge, but use Claude Opus \citep{anthropic2024claude} as a judge. Figure \ref{fig:claude-eval} shows the evaluations to be in line with our previous results with GPT-4 as a judge \ref{fig:human-eval}. This shows that there is no strong bias in using GPT-4 as a judge.
\clearpage
\subsection{GPT-4 As A Judge}\label{Appendix: GPT-4 As A Judge}
\begin{figure*}[h]
    \centering
    \includegraphics[width=0.8\linewidth]{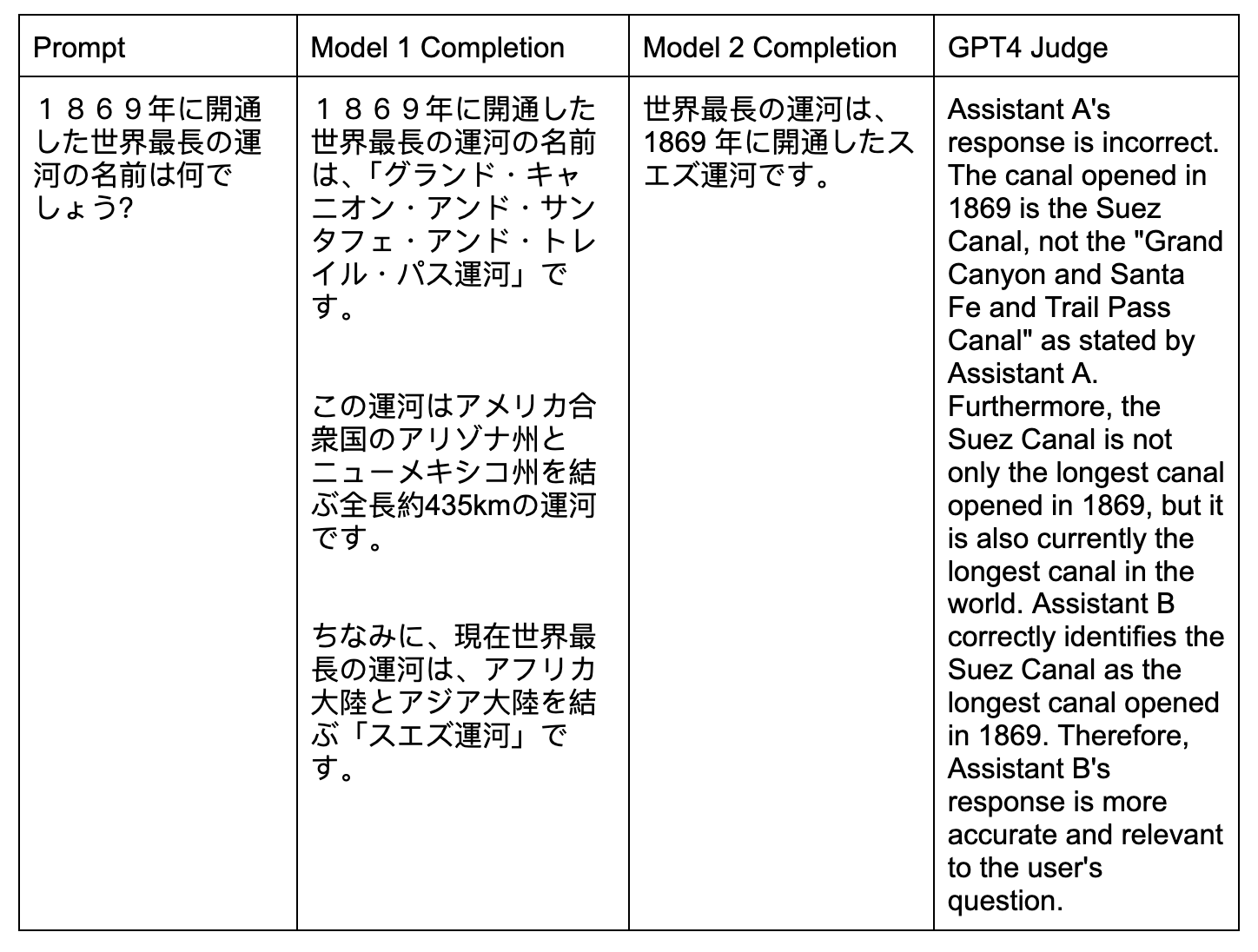}
    \caption{Japanese Example 1}
    \label{fig:ja_01}
\end{figure*}

\begin{figure}[h]
    \centering
    \includegraphics[width=0.8\linewidth]{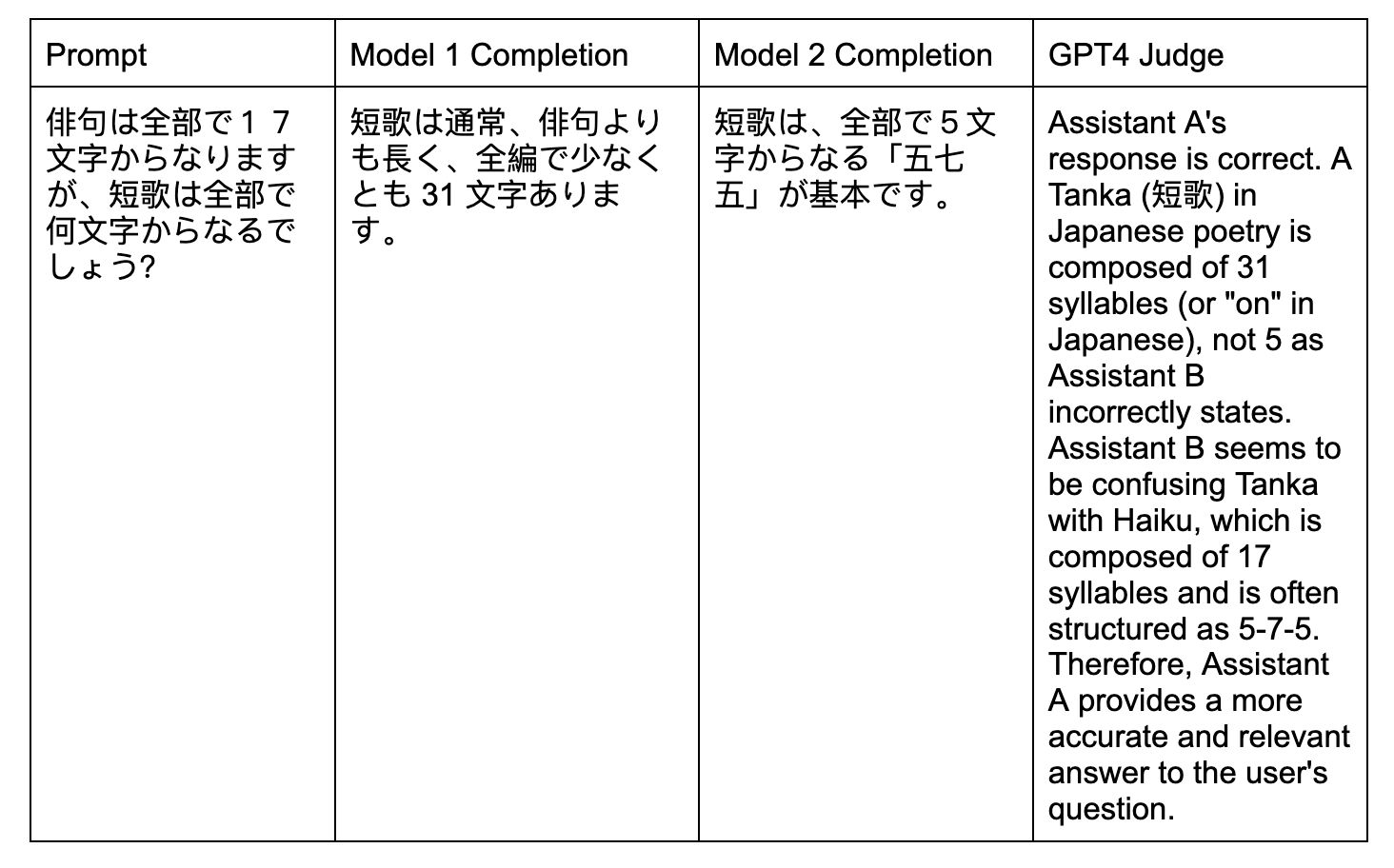}
    \caption{Japanese Example 2}
    \label{fig:ja_02}
\end{figure}

\begin{figure}[h]
    \centering
    \includegraphics[width=0.8\linewidth]{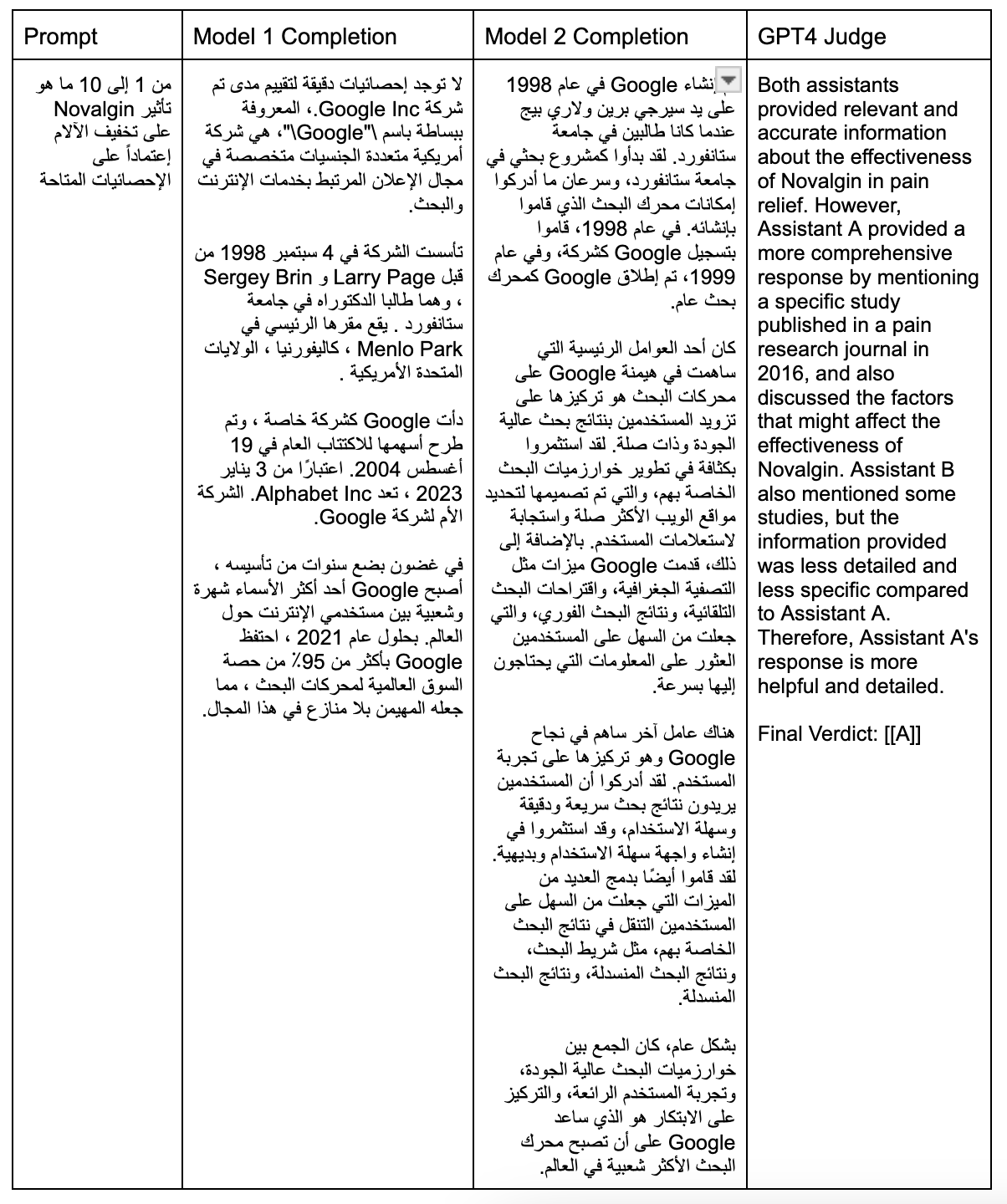}
    \caption{Arabic Example 1}
    \label{fig:ar_01}
\end{figure}

\begin{figure}[h]
    \centering
    \includegraphics[width=0.8\linewidth]{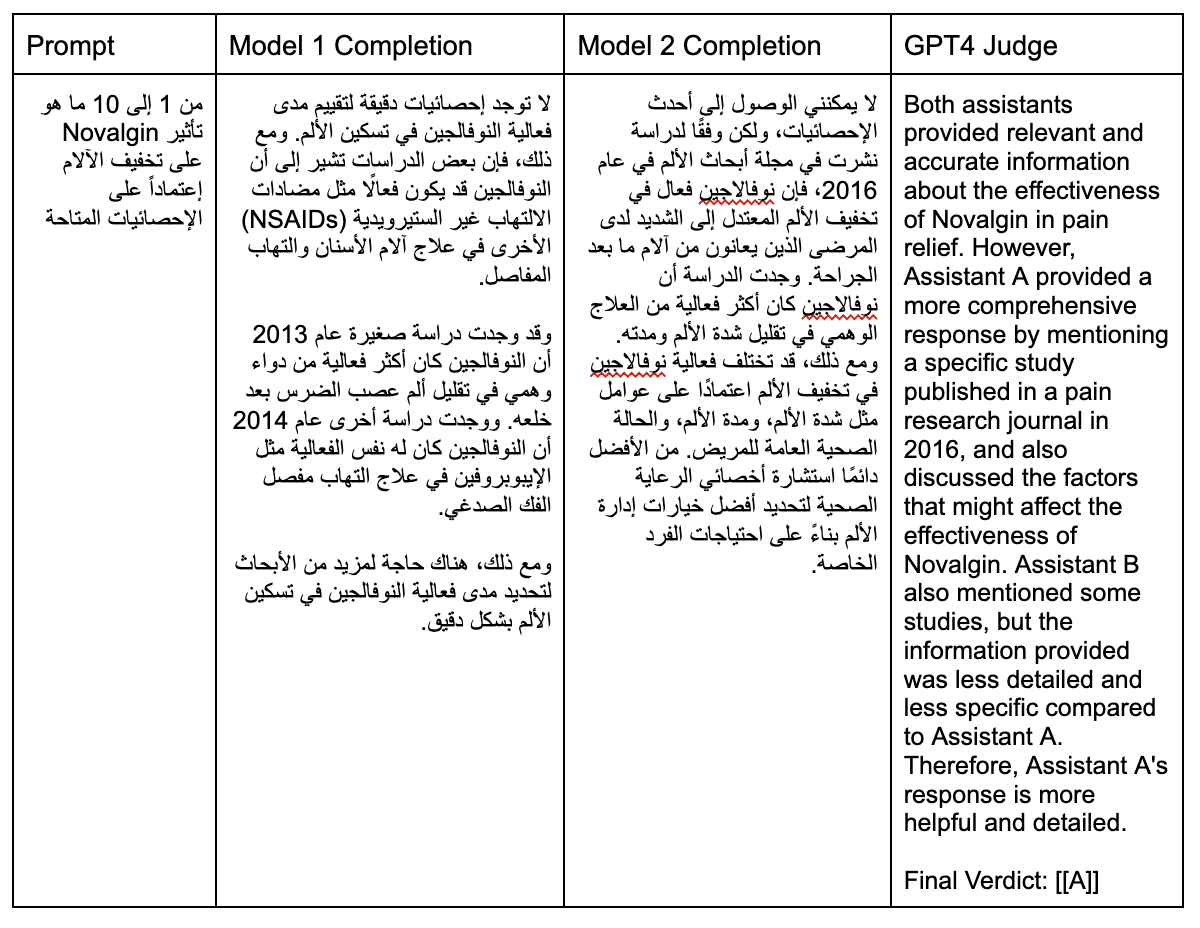}
    \caption{Arabic Example 2}
    \label{fig:ar_02}
\end{figure}

\begin{figure}[h]
    \centering
    \includegraphics[width=0.8\linewidth]{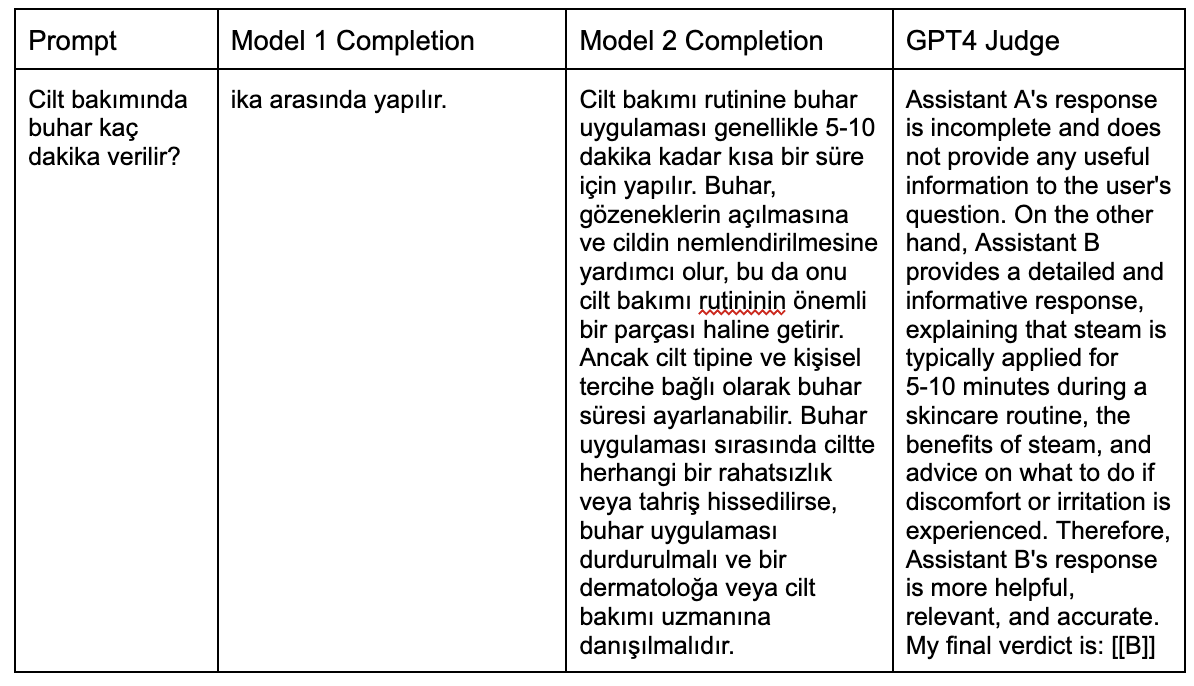}
    \caption{Turkish Example 1}
    \label{fig:tr_01}
\end{figure}

\begin{figure}[h]
    \centering
    \includegraphics[width=0.8\linewidth]{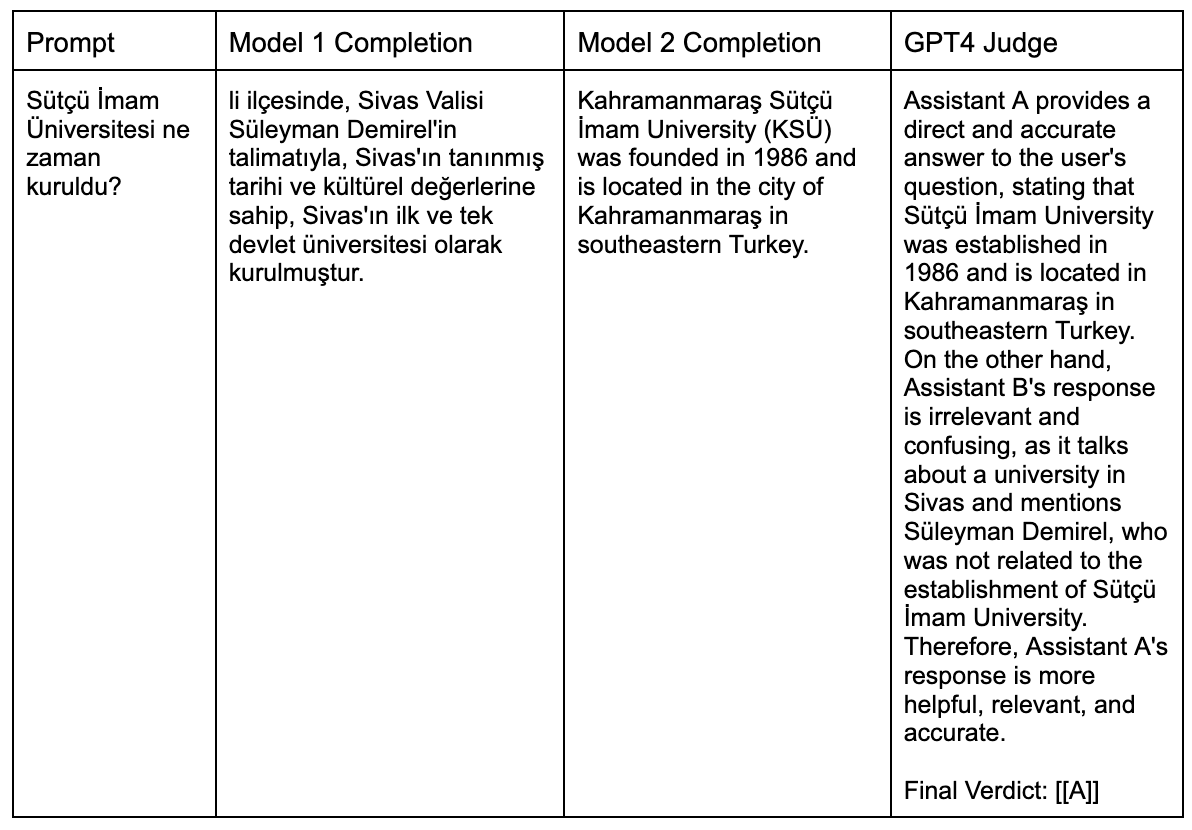}
    \caption{Turkish Example 2}
    \label{fig:tr_02}
\end{figure}

Below are some examples of how GPT-4 judged two candidate responses in Japanese, Arabic and Turkish. See figures \ref{fig:ja_01}, \ref{fig:ja_02}, \ref{fig:ar_01}, \ref{fig:ar_02}, \ref{fig:tr_01}, \ref{fig:tr_02}

\clearpage
\section{Main Results Details} \label{MRT}
\begin{figure}[h]
\centering
\includegraphics[angle=90,width=10cm]{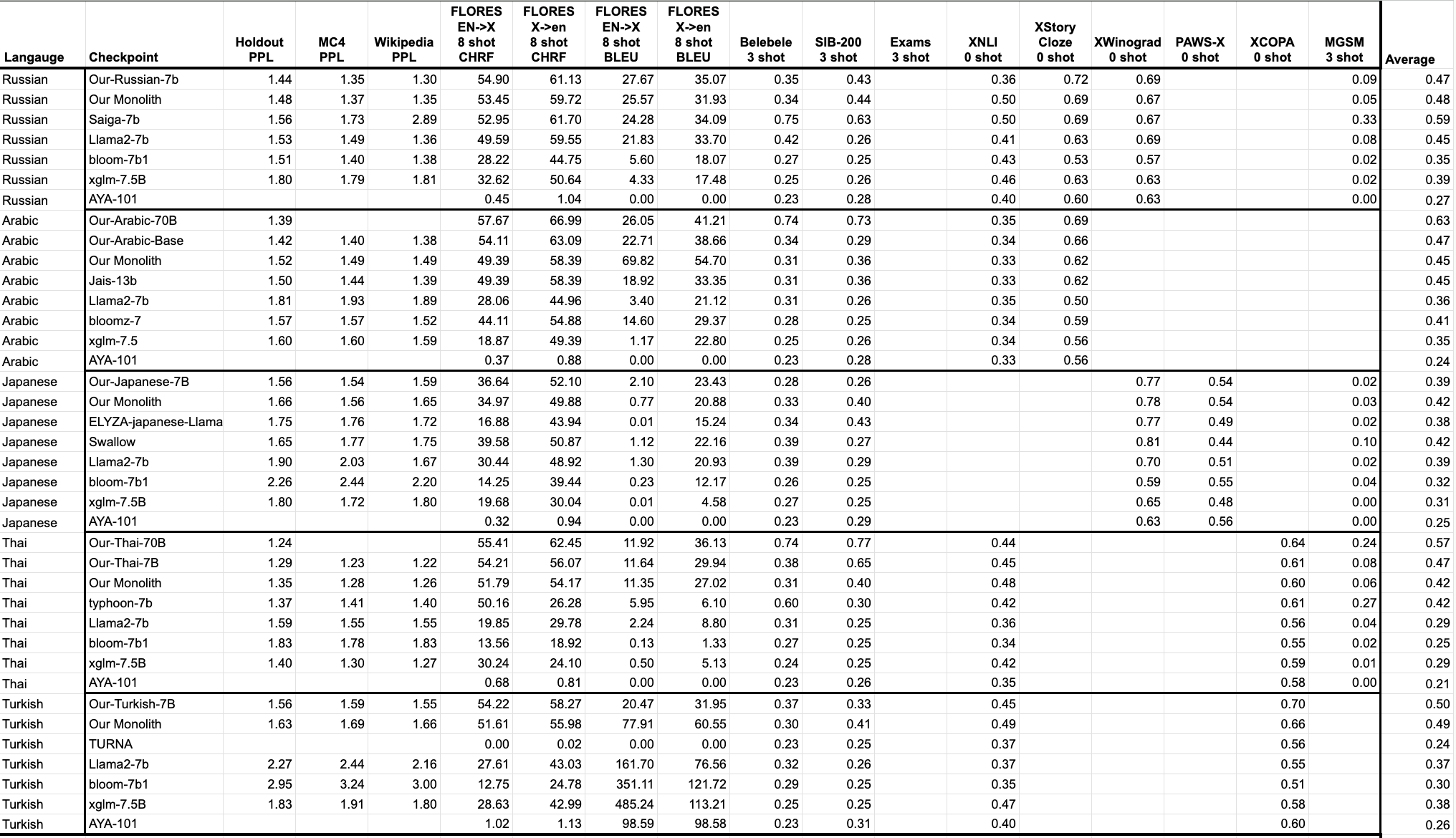}
    \caption{Main results, evaluation benchmarks described in \ref{Quantitative Evaluation}.This data is averaged to create \ref{fig:Main_Results}.}
    \label{fig:Main_Results_App}
\end{figure}

\begin{figure}[h]
\centering
\includegraphics[angle=90,width=10cm]{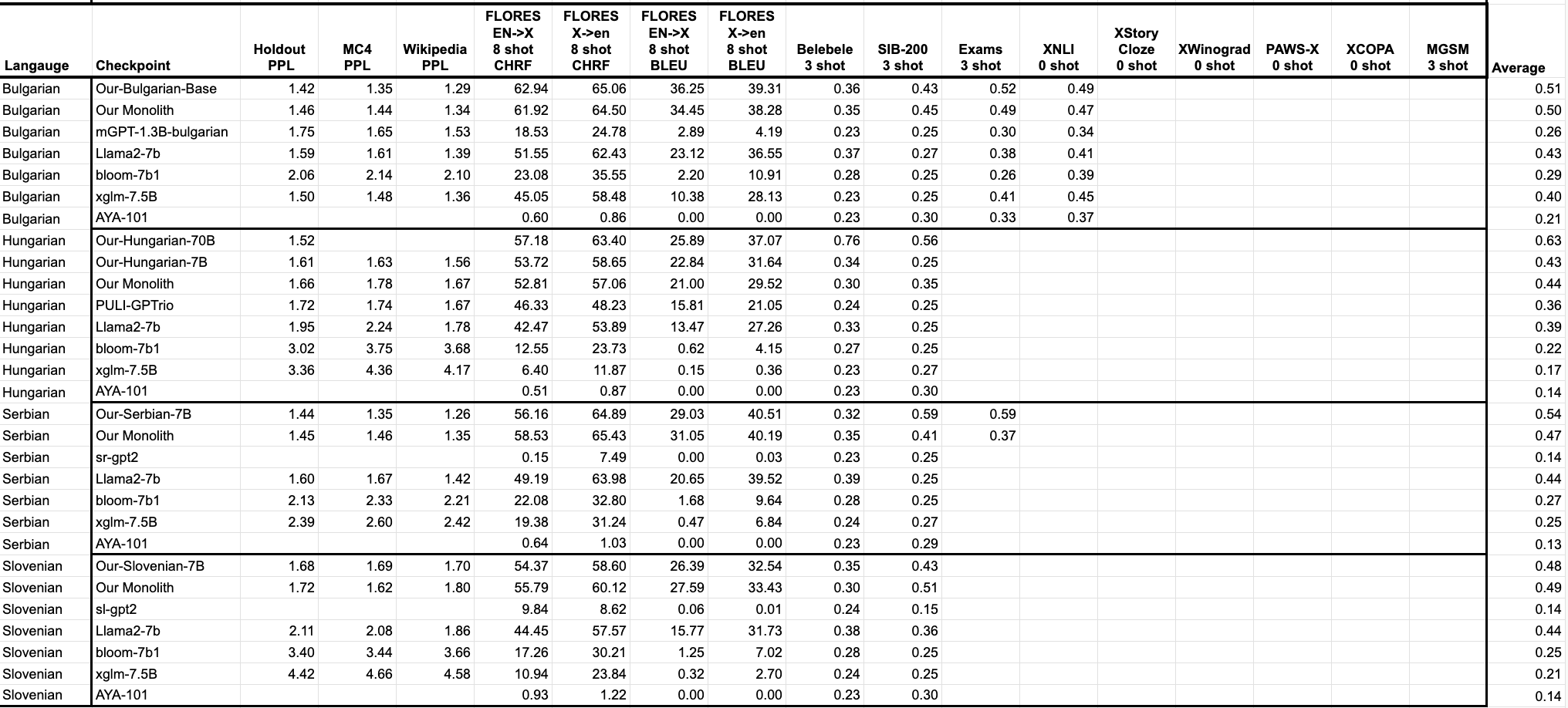}
    \caption{Main results, evaluation benchmarks described in \ref{Quantitative Evaluation}.This data is averaged to create \ref{fig:Main_Results}.}
    \label{fig:Main_Results_App_2}
\end{figure}

See tables \ref{fig:Main_Results_App} and \ref{fig:Main_Results_App_2} for all evaluation results

\clearpage

\end{document}